\pdfoutput=1

\documentclass[11pt]{article}

\usepackage{ACL2023}

\usepackage{times}
\usepackage{ulem}
\usepackage{graphicx}
\usepackage{latexsym}
\usepackage{multirow}
\usepackage{amssymb}
\usepackage{amsmath}
\usepackage{booktabs}
\usepackage{dsfont}
\usepackage[T1]{fontenc}

\usepackage[utf8]{inputenc}
\usepackage{url}

\usepackage{microtype}

\usepackage{inconsolata}

\title{PCRED: Zero-shot Relation Triplet Extraction with Potential Candidate Relation Selection and Entity Boundary Detection}

\author{Yuquan Lan$^1$, Dongxu Li$^1$, Yunqi Zhang$^1$, Hui Zhao$^{1,2}\thanks{~~Corresponding Author}$, Gang Zhao$^3$\\
  $^{1}$Software Engineering Institute, East China Normal University, Shanghai, China \\
  $^{2}$Shanghai Key Laboratory of Trustworthy Computing, Shanghai, China \\
  $^{3}$Microsoft, China \\
  \texttt{\{yqlan,lidx,yunqi.zhang\}@stu.ecnu.edu.cn} \\
  \texttt{hzhao@sei.ecnu.edu.cn} \\
  \texttt{gang.zhao@microsoft.com}
}
\begin{document}
\maketitle
\begin{abstract}
Zero-shot relation triplet extraction (ZeroRTE) aims to extract relation triplets from unstructured texts under the zero-shot setting, where the relation sets at the training and testing stages are disjoint. 
Previous state-of-the-art method handles this challenging task by leveraging pretrained language models to generate data as additional training samples, which increases the training cost and severely constrains the model performance.
To address the above issues, we propose a novel method named PCRED for ZeroRTE with \textbf{P}otential \textbf{C}andidate \textbf{R}elation Selection and \textbf{E}ntity Boundary \textbf{D}etection. 
The remarkable characteristic of PCRED is that it does not rely on additional data and still achieves promising performance.
The model adopts a relation-first paradigm, recognizing unseen relations through candidate relation selection. 
With this approach, the semantics of relations are naturally infused in the context.
Entities are extracted based on the context and the semantics of relations subsequently.
We evaluate our model on two ZeroRTE datasets.
The experiment results show that our method consistently outperforms previous works.
Our code will be available at \url{https://anonymous.4open.science/r/PCRED}.
\end{abstract}
\section{Introduction}
Relation triplet extraction (RTE) is a crucial task in information extraction, which aims to extract relation triplets from unstructured text. 
It has wide applications, such as knowledge base enrichment~\cite{trisedya-etal-2019-neural} and question-answering system construction~\cite{xu2016question}.

There have been various methods for RTE.
The paradigm ranges from tagging-based methods~\citep{wei-etal-2020-novel,wang-etal-2020-tplinker}, span-based methods~\cite{zhong-chen-2021-frustratingly}, table-based methods~\citep{ren-etal-2021-novel,ma-etal-2022-joint} to sequence-to-sequence methods~\cite{huguet-cabot-navigli-2021-rebel-relation}.
However, these models can only extract triplets whose relations are included in the predefined relation set. 
Consequently, the applications of these models in real-world scenarios are limited.

\begin{figure}
	\centering
	\includegraphics[width=\linewidth]{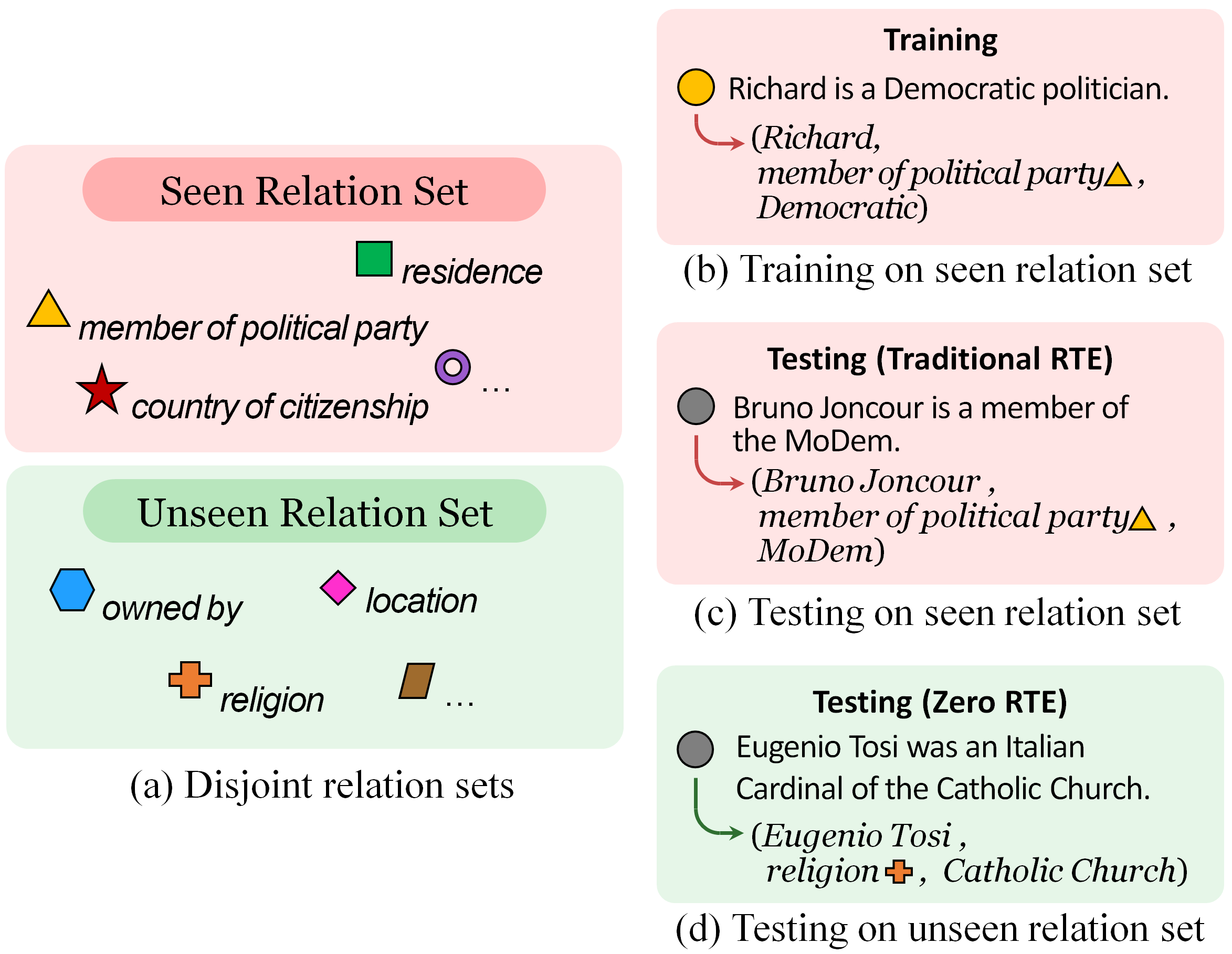}
	\caption{An example to illustrate the difference between traditional RTE and ZeroRTE}
	\label{fig:example}
\end{figure}

To this end, \citet{chia-etal-2022-relationprompt} propose a challenging task called zero-shot relation triplet extraction (ZeroRTE) to overcome the above limitations and encourage further research on training models that can generalize to unseen relations.
ZeroRTE aims to extract relational triplets under the zero-shot setting.
Figure~\ref{fig:example} provides an example to better illustrate the difference between traditional RTE and ZeroRTE.
The relation sets at the training and testing stages are disjoint (Figure~\ref{fig:example}(a)). 
Models for ZeroRTE are trained on the seen relation set and extract triplets with unseen relations at the testing stage (Figure~\ref{fig:example}(b, d)), while models for traditional RTE are only required to extract triplets with the same seen relations (Figure~\ref{fig:example}(b, c)).

Due to the challenging task setting, few previous RTE methods can be applied to ZeroRTE.
For instance, previous methods like PRGC~\cite{zheng-etal-2021-prgc} and GRTE~\cite{ren-etal-2021-novel}
treat relations as either discrete class IDs or learnable embeddings. OneRel~\cite{shang2022onerel} directly tags relations in a three-dimensional matrix. 
Since both the relation type and the relation number are different between training and testing, these methods can hardly solve the ZeroRTE task.
REBEL~\cite{huguet-cabot-navigli-2021-rebel-relation}, a typical sequence-to-sequence method, employs BART~\cite{lewis-etal-2020-bart} as the backbone model to generate formatted sequences that can be decoded into triplets. 
However, it is unlikely to generate sequences that contain unseen relations.

Previous state-of-the-art method RelationPrompt~\cite{chia-etal-2022-relationprompt} extends the sequence-to-sequence method to ZeroRTE.
The core idea is to leverage structured templates to prompt pretrained language models (PLMs) like GPT2~\cite{radford2019language} to generate synthetic data as additional training samples of unseen relations.
The synthetic data help to train another model to extract triplets. 
Although utilizing PLMs to generate training data significantly saves the cost of human annotation,
it raises new problems.
The model performance severely depends on the synthetic data, without which it degrades dramatically.
Besides, it is challenging to control the quality of the generated data.

In this paper, we propose a novel method named PCRED with potential candidate relation selection and entity boundary detection.
One remarkable characteristic of PCRED is that it does not rely on any additional training samples and still achieves better performances.
PCRED adopts a relation-first paradigm for triplet extraction based on the fact that relations are usually triggered by context rather than entities~\cite{li-etal-2022-rfbfn}.
The semantics of relations are infused by concatenating relations to the original sentences as additional context. The triplet extraction is progressed with three steps: potential candidate relation selection, relation filtering and entity boundary detection.
We first aggregate a group of candidate relations and train the model to select potential relations expressed in the sentence.
Through relation selection rather than relation classification, the model can recognize unseen relations based on their semantics.
Relations whose probability is lower than the threshold are removed as irrelevant relations.
Afterward, entity extraction is performed by recognizing entity boundaries according to the specific relation.

Compared with RelationPrompt, our PCRED has several advantages.
First, our model does not rely on synthetic data, which saves the cost of training a generator for synthetic data generation.
Besides, we only employ the base version of BERT ~\cite{devlin-etal-2019-bert} as the backbone model, which has approximately 150M parameters. RelationPrompt comprises a relation generator and a triplet extractor, which has 264M parameters (124M for the generator and 140M for the extractor). Our model has only 56.8\% of the parameters against to RelationPrompt.

We perform comprehensive experiments on two ZeroRTE datasets. 
The experiment results show that our proposed method consistently outperforms previous works.  
The main contributions of this paper are summarized as follows:
\begin{enumerate}
	\item We propose a novel method named PCRED with potential candidate relation selection and entity boundary detection, which consistently outperforms previous works on two datasets.
 
	\item We tackle the ZeroRTE task from a new perspective. 
	Instead of leveraging PLMs to generate training samples of unseen relations, our model directly utilizes the semantics of unseen relations, hence no additional data and training cost.
	
	\item Our proposed PCRED recognizes unseen relations by potential candidate relation selection, which naturally fits the zero-shot setting and infuses the semantics of relations in the context. To the best of our knowledge, this is the first exploration of ZeroRTE.
\end{enumerate}
\section{Methodology}
\begin{figure*}
	\centering
	\includegraphics[width=\linewidth]{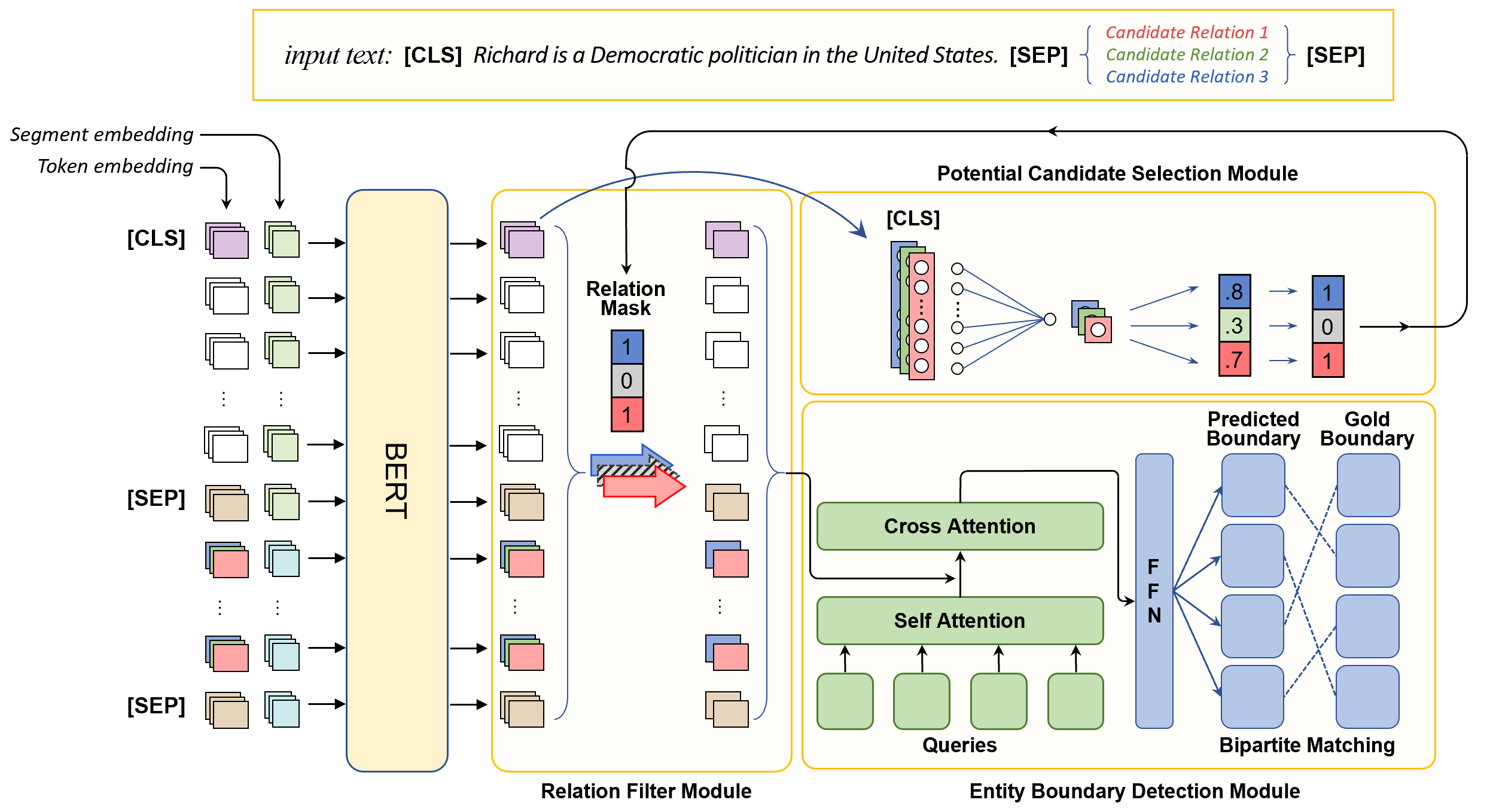}
	\caption{The model architecture}
	\label{fig:model}
\end{figure*}
\subsection{Task Formulation}
Let $\mathcal{D}=(\mathcal{S},\mathcal{T},\mathcal{Y})$ denotes the whole dataset, consisting of the input sentences $\mathcal{S}$, the output triplets $\mathcal{T}$ and the set of relation labels $\mathcal{Y}$. 
$\mathcal{D} = \mathcal{D}_s \cup \mathcal{D}_u$, where $\mathcal{D}_s,\mathcal{D}_u$ refer to the seen and unseen datasets respectively. 
The model is trained on $\mathcal{D}_s$ and evaluated on $\mathcal{D}_u$.
$\mathcal{Y} = \mathcal{Y}_s \cup \mathcal{Y}_u $ is predefined, comprising the seen relation label set $\mathcal{Y}_s=\{y_s^1,\cdots, y_s^n\}$ and unseen relation label set $\mathcal{Y}_u=\{y_u^1,\cdots, y_u^m\}$, where $n = |\mathcal{Y}_s|$ and $m = |\mathcal{Y}_u|$ are the number of relation labels. $\mathcal{Y}_s$ and $\mathcal{Y}_u$ are disjoint, $\mathcal{Y}_s \cap \mathcal{Y}_u = \varnothing$. 
One sentence $s \in \mathcal{S}$ contains one or more triplets. 
A triplet $t \in \mathcal{T}$ is defined as $(e_{head}, e_{tail}, y)$, where $e_{head}, e_{tail}$ refer to the head and tail entity, and $y\in \mathcal{Y}$ is the relation that holds between these two entities.
The model takes the sentence $s$ as input and outputs a list of triplets.  
\subsection{Model Architecture}
Figure~\ref{fig:model} demonstrates the overview model architecture of our proposed RCRED.
BERT is employed as the sentence encoder to get the contextual representation.
PCRED comprises three modules, the potential candidate relation selection module, the relation filter module, and the entity boundary detection module. They are introduced in the following subsections.
\begin{table*}[h!]
	\centering
	\begin{tabular}{|r|p{9cm}|}
	\hline
	\bfseries{Context} & Richard is a Democratic politician in the United States. \\
	\hline
	\multirow{2}{*}{\textbf{Triplets}} & (Richard, member of political party, Democratic politician) \\
	                    & (Richard, country of citizenship, the United States) \\
	\hline
	\multirow{3}{*}{\textbf{Candidate Relations}} & member of political party\\
	& position held \\
	& country of citizenship\\
	\hline
	\multirow{6}{*}{\textbf{Augmented Sentences}} & [CLS] Richard is a Democratic politician in the United States. [SEP] member of political party [SEP] \\
	& [CLS] Richard is a Democratic politician in the United States. [SEP] position held [SEP] \\
	& [CLS] Richard is a Democratic politician in the United States. [SEP] country of citizenship [SEP] \\
	\hline
	\bfseries{Potential Relations} & member of political party, country of citizenship \\
	\hline
	\bfseries{Relation Mask} & [1, 0, 1] \\
	\hline
	\multirow{2}{*}{\bfseries{Filtered Sentences}} & \sout{[CLS] Richard is a Democratic politician in the United States. [SEP] position held [SEP]} \\
	\hline
	\end{tabular}
	\caption{An example of the potential candidate relation selection and relation filtering}
	\label{tab:example}
\end{table*}
\subsubsection{Potential Candidate Relation Selection Module}
The potential candidate relation selection module recognizes relations expressed in the context.
Since relations at the testing stage are unseen, we model this task as a relation selection problem.

Assume there are at most $\mathcal{G}$ different relations in a sentence. 
We expand the input sentence $s$ into $\mathcal{G}$ copies, $\{s_i\}_{i=1}^\mathcal{G}$.
For each copy, we denote the text of each relation $y_i,i\in\{1,...,\mathcal{G}\}$ as $r_i$ and concatenate them to the original sentence to get the augmented sentences $s'$:
\begin{equation}
    s'=\{\text{[CLS]}~s_i~\text{[SEP]}~r_i~\text{[SEP]}\}_{i=1}^\mathcal{G}
\end{equation}
where [CLS] and [SEP] are two special markers used in BERT.
The original sentence and the relation are distinguished with different segment ids.
With this approach, the semantics of relations are naturally infused in the sentence as additional context. 

The augmented sentences $s'$ are tokenized and mapped to ids as $X$, which is fed into BERT to get the contextual representation $H$:
\begin{equation}
    H = \text{BERT}(X)
\end{equation}
where $H \in \mathbb{R}^{\mathcal{G} \times l \times d}$, $l$ denotes the sequence length, and $d$ is the hidden dimension of BERT.

The representation of the special token [CLS] $H_c \in \mathbb{R}^{\mathcal{G} \times d}$ is passed to a fully connected layer and activated by tanh function to get the pooling representation $H_p \in \mathbb{R}^{\mathcal{G} \times d}$:
\begin{equation}
    H_p = \tanh(H_c\cdot W_p^\top + b_p)
\end{equation}
where $W_p \in \mathbb{R}^{d \times d}$ and $b_p$ are the learnable weight matrix and bias. 

$H_p$ is fed into the sigmoid function to get the probability distributions over $\mathcal{G}$ candidate relations:
\begin{equation}
    P_r = \text{sigmoid}(H_{[pooling]}\cdot W_c^\top + b_c)
\end{equation}
where $W_c \in \mathbb{R}^{d \times 1}$ and $b_c$ are the learnable weight matrix and bias. 

The value of $\mathcal{G}$ can be different on the training, validation and testing sets.
At the training stage, we set $\mathcal{G}$ as the maximum number of different relations expressed in a sentence.
If one sentence has only $\mathcal{K}~(\mathcal{K}<\mathcal{G})$ different relations, we randomly sample $\mathcal{G}-\mathcal{K}$ relations as irrelevant relations.
We aggregate the ground-truth relations and the sampled irrelevant relations to get $\mathcal{G}$ candidate relations.
At the validation and testing stages, since the ground-truth relations are invisible, we treat all the relations in the set as candidate relations. In other words, $\mathcal{G}$ is equal to $|\mathcal{Y}_u|$.

\subsubsection{Relation Filter Module}
The relation filter module filters out irrelevant relations using a boolean relation mask.
At the training stage, we use the ground-truth relation labels as the relation mask;
at the validation and testing stages, the relation mask is defined as:
\begin{equation}
    M_i = \begin{cases} 1 & P_r \ge \delta \\ 0 &  P_r < \delta\end{cases}
\end{equation}
where $\delta$ is the relation threshold.

The contextual representation $H\in \mathbb{R}^{\mathcal{G} \times l \times d}$ is fed into the relation filter module and transformed into the filtered representation $H_f \in \mathbb{R}^{\lambda \times l \times d} $:
\begin{equation}
    H_f = \text{RelationFilter}(H, M)
\end{equation}
where $\lambda \le \mathcal{G}$ is the number of potential relations.

We provide an example in Table~\ref{tab:example} to better illustrate the process. Assume $\mathcal{G}=3$ and there are 3 candidate relations. The relation \textit{position held} is an irrelevant relation and will be filtered out by the relation filter module (marked with a strikethrough).
\subsubsection{Entity Boundary Detection Module}
Since entities are continuous spans of tokens, they can be determined by the start and end positions in the sentence.
Based on the semantics of relations infused in the context, the entity boundary detection module aims to detect the entity boundary, which consists of four indexes $\{h^{start}, h^{end}, t^{start}, t^{end}\}$ denoting the start and end positions of the head entity and tail entity respectively.
Since one sentence may contain multiple entities, the module integrates a set prediction network to jointly extract multiple boundaries, which is inspired by SPN~\cite{sui2020joint}.
Since SPN treats relations as discrete IDs, it ignores the semantics of relations.
Consequently, SPN is not applicable to ZeroRTE.

We first randomly initialize $\mathcal{N}$ trainable embeddings $Q \in \mathbb{R}^{\mathcal{N} \times d}$, where $\mathcal{N}$ is the maximum number of triplets per relation. 
$Q$ is first fed to a self-attention layer, yielding a hidden representation $\mathcal{Q} \in \mathbb{R}^{\mathcal{N} \times d}$.
$\mathcal{Q}$ and $H_f$ are fed into a cross-attention layer to get the output representation $H_o \in \mathbb{R}^{\mathcal{N} \times d}$, which models the association between entities and semantics of relations.
Four linear layers are used to predict the $j$th boundary of entities for each selected relation:
\begin{gather}
    P^{h^{start}}_j = \text{softmax(FFN(GELU}(h_jW_1 + H_fW_2))) \\
    P^{h^{end}}_j = \text{softmax(FFN(GELU}(h_jW_3 + H_fW_4)))   \\
    P^{t^{start}}_j = \text{softmax(FFN(GELU}(h_jW_5 + H_fW_6)))  \\
    P^{t^{end}}_j = \text{softmax(FFN(GELU}(h_jW_7 + H_fW_8)))
\end{gather}
where FFN is the feed forward network, GELU is the gaussian error linear units function, $h_j \in \mathbb{R}^{d}$ is the $j$th representation of $H_o$, and $\{W_k \in \mathbb{R}^{d \times d}\}_{k=1}^8$ are learnable parameters.
\subsection{Training}
\begin{table*}[ht]
	\centering
	\begin{tabular}{cccccccc}
		\toprule
		& Samples & Entities & \multicolumn{4}{c}{Relation Labels} & Average Length \\
		\midrule
		& & & Total & Train & Validation & Test & \\
		\midrule
		\multirow{3}{*}{Wiki-ZSL} & \multirow{3}{*}{94383} & \multirow{3}{*}{77623} & \multirow{3}{*}{113} & 103 & 5 & 5 & \multirow{3}{*}{24.85} \\
		& & & & 98 & 5 & 10 & \\
		& & & & 93 & 5 & 15 & \\
		\midrule
		\multirow{3}{*}{FewRel} & \multirow{3}{*}{56000} & \multirow{3}{*}{72954} & \multirow{3}{*}{80} & 70 & 5 & 5 & \multirow{3}{*}{24.95} \\
		& & & & 65 & 5 & 10 & \\
		& & & & 60 & 5 & 15 & \\
		\bottomrule
	\end{tabular}
	\caption{Statistics of FewRel and Wiki-ZSL}
	\label{tab:statistics}
\end{table*}
There are two training tasks in our model: potential candidate relation selection and entity boundary detection.
We train the model in a multi-task manner.
For potential candidate relation selection, the training objective is to minimize the binary cross entropy loss:
\begin{equation}
    \mathcal{L}_{rel} = -\frac{1}{\mathcal{G}}\sum_{i=1}^\mathcal{G} (y_i \log P_{r} + (1-y_i) \log (1 - P_{r}))  
\end{equation}
For entity boundary detection, we adopt the bipartite matching loss~\cite{sui2020joint}, which is invariant to any permutation of predictions.
We denote the $j$th tuple of the ground-truth entity boundary set as $Y_j=(h_j^{start}, h_j^{end}, t_j^{start}, t_j^{end})$, and the $j$th tuple of the predicted entity boundary set as $\hat{Y}_j=(P^{h^{start}}_j,P^{h^{end}}_j,P^{t^{start}}_j,P^{t^{end}}_j)$.
To find an optimal matching between the set of ground-truth $Y$ and the set of prediction $\hat{Y}$, we apply the Hungarian algorithm\footnote{\url{https://en.wikipedia.org/wiki/Hungarian_algorithm}} to search for a permutation of elements $\pi^{*}$ with the lowest cost:
\begin{equation}
    \pi^{*} = \underset{\pi \in\prod(\mathcal{N})}{\text{argmin}}
    \sum_{j=1}^\mathcal{N} \mathcal{C}_{match}(Y_j, \hat{Y}_{\pi(j)})
\end{equation}
where $\prod(\mathcal{N})$ is the space of permutations, and $\mathcal{C}_{match}(Y_j, \hat{Y}_{\pi(j)})$ is a pair-wise matching cost defined as:
\begin{equation}
\begin{aligned}
    \mathcal{C}_{match}(Y_j, \hat{Y}_{\pi(j)}) &= -\mathds{1}[P_{\pi(j)}^{h^{start}}(h_j^{start}) \\
    &+ P_{\pi(j)}^{h^{end}}(h_j^{end}) \\
    &+ P_{\pi(j)}^{t^{start}}(t_j^{start}) \\
    &+ P_{\pi(j)}^{t^{end}}(t_j^{end})]
\end{aligned}
\end{equation}

The entity loss is defined as:
\begin{equation}
\begin{aligned}
    \mathcal{L}_{ent} &= -\frac{1}{\mathcal{N}}\sum_{j=1}^{\mathcal{N}}\mathds{1}[\log P_{\pi^*(j)}^{h^{start}}(h_j^{start}) \\
    &+ \log P_{\pi^*(j)}^{h^{end}}(h_j^{end}) \\
    &+ \log P_{\pi^*(j)}^{t^{start}}(t_j^{start}) \\
    &+ \log P_{\pi^*(j)}^{t^{end}}(t_j^{end})]
\end{aligned}
\end{equation}

The overall loss is calculated as:
\begin{equation}
    \mathcal{L} = \alpha \mathcal{L}_{rel} + (1-\alpha) \mathcal{L}_{ent}
\end{equation}
where $\alpha$ controls the weight between these two training objectives.
\subsection{Inference}
To extract the triplets, the model first performs potential candidate relation selection and filters out irrelevant ones.
For each potential relation $y_i,i\in \{1,\cdots, \mathcal{G}\}$, the entity boundary module outputs $\mathcal{N}$ possible boundaries, $\{(h_j^{start},h_j^{end},t_j^{start},t_j^{end})\}_{j=1}^\mathcal{N}$. 
We remove invalid entity boundaries that violate the following criteria:
\begin{itemize}
    \item The start index should be smaller than or equal to the end index.
    \item The end index should not exceed the actual length of the sentence.
    \item The end index minus the start index should not exceed the maximum span length.
    \item The probability of the extracted boundary should be higher than a threshold, i.e.,
    $P^{h^{start}} \times P^{h^{end}} \times P^{t^{start}} \times P^{t^{end}} \ge \beta$, where $\beta$ is the boundary threshold.
\end{itemize}

Each selected relation $y_i$ and the filtered entity boundaries $\{(h_j^{start},h_j^{end},t_j^{start},t_j^{end})\}_{j=1}^{\mathcal{N}_f}~(\mathcal{N}_f\le \mathcal{N})$ are combined to constitute the final triplets.
\section{Experiment}
\subsection{Dataset}
We evaluate our model on two ZeroRC datasets - Wiki-ZSL~\cite{chen-li-2021-zs} and FewRel~\cite{han-etal-2018-fewrel}.
Wiki-ZSL is constructed by distant supervision over Wikipedia articles.
FewRel is originally designed for few-shot relation classification.
~\citet{chia-etal-2022-relationprompt} transform it for ZeroRTE by splitting data into disjoint relation sets for training, validation, and testing.
They randomly selected $m$ relations as the unseen relations for validation and testing. 
The remaining ones are treated as seen relations for training.
The selection process is repeated 5 times with 5 different random seeds, producing 5 data folds.
In the validation set, $m$ is fixed to 5.
In the test set, $m$ has three different settings, $m\in\{5,10,15\}$, which aims to study the model performance under different numbers of unseen relation labels.
More statistics of these two datasets are shown in Table~\ref{tab:statistics}. 
We use the same processed datasets provided by~\citet{chia-etal-2022-relationprompt} to conduct the experiments. 
\begin{table*}[h!]
	\centering
	\resizebox{\linewidth}{!}{
	\begin{tabular}{cccccccccc}
		\toprule
		& &\multicolumn{2}{c}{Single Triplet} & \multicolumn{6}{c}{Multi Triplet} \\
		\bfseries Labels & \bfseries  Model & \bfseries Wiki-ZSL & \bfseries FewRel &  \multicolumn{3}{c}{\bfseries Wiki-ZSL} &  \multicolumn{3}{c}{\bfseries FewRel} \\
		& & $Acc$. & $Acc.$ & $Pre.$ & $Rec.$ & $F_1$ & $Pre.$ & $Rec.$ & $F_1$ \\
		\midrule
		\multirow{4}{*}{m=5}&TableSequence & 14.47 & 11.82 & \textbf{43.68} & 3.51 & 6.29 & 15.23 & 1.91 & 3.40 \\
		&RelationPrompt (NoGen) & 9.05 & 11.49 & 15.58 & \textbf{43.23} & 22.26 & 9.45 & \textbf{36.74} & 14.57 \\
		&RelationPrompt & 16.64 & 22.27 & 29.11 & 31.00 & 30.01 & 20.80 & 24.32 & 22.34 \\
		&PCRED (Ours) & \textbf{18.40} & \textbf{22.67} & 38.14 & 36.84 & \textbf{37.48} & \textbf{43.91} & 34.97 & \textbf{38.93} \\
		\midrule 
		\multirow{4}{*}{m=10}&TableSequence & 9.61 & 12.54 & \textbf{45.31} & 3.57 & 6.4 & 28.93 & 3.60 & 6.37 \\
		&RelationPrompt (NoGen) & 7.10 & 12.40 & 9.63 & \textbf{45.01} & 15.70 & 6.40 & \textbf{41.70} & 11.02 \\
		&RelationPrompt & 16.48 & 23.18 & 30.20 & 32.31 & 31.19 & 21.59 & 28.68 & 24.61 \\
		&PCRED (Ours) & \textbf{22.30} & \textbf{24.91} & 27.09 & 39.09 & \textbf{32.00} & \textbf{30.89} & 29.90 & \textbf{30.39} \\
		\midrule    
		\multirow{4}{*}{m=15}&TableSequence & 9.20 & 11.65 & \textbf{44.43} & 3.53 & 6.39 & 19.03 & 1.99 & 3.48\\
		&RelationPrompt (NoGen) & 6.61 & 10.93 & 7.25 & \textbf{44.68} & 12.34 & 4.61 & \textbf{36.39} & 8.15 \\
		&RelationPrompt & 16.16 & 18.97 & 26.19 & 32.12 & 28.85 & 17.73 & 23.20 & 20.08 \\
		&PCRED (Ours) & \textbf{21.64} & \textbf{25.14} & 25.37 & 33.80 & \textbf{28.98} & \textbf{27.00} & 23.55 & \textbf{25.16} \\
		\bottomrule
	\end{tabular}
	}
	\caption{Main Results.}
	\label{tab:result}
\end{table*}

\subsection{Evaluation Metrics}
To evaluate the performance of our model, we follow the same evaluation metrics as RelationPrompt~\cite{chia-etal-2022-relationprompt} for a fair comparison.
We separately report the scores for sentences that contain a single triplet and multiple triplets to be consistent with previous studies.
For single triplet extraction, we use Accuracy ($Acc.$) as the evaluation metric; for multiple triplet extraction, we use Micro-F1 score ($F_1$) as the evaluation metric and report the precision ($Pre.$) as well as the recall ($Rec.$) score.
All the scores are the average results across five data folds.
More details about implementation are listed in Appendix~\ref{sec:appendix}.
\subsection{Compared Methods}
Since ZeroRTE is a challenging new task, few previous traditional RTE methods can be applied to ZeroRTE.
We compare PCRED with methods presented by~\citet{chia-etal-2022-relationprompt}.

RelationPrompt~\cite{chia-etal-2022-relationprompt} is the previous state-of-the-art method, which comprises a relation generator and a triplet extractor.
The relation generator is based on GPT2, which generates synthetic data of unseen relations to train the triplet extractor.
The triplet extractor is based on BART, which generates sequences to be decoded into triplets.
The model denoted as ``NoGen'' means that it directly uses the triplet extractor for extraction without synthetic data.

TableSequence~\cite{wang-lu-2020-two} is a typical table-based method, which comprises two encoders to encode different types of information in the learning process. Since it cannot directly tackle ZeroRTE,~\citet{chia-etal-2022-relationprompt} provide the synthetic data generated from the relation generator of RelationPrompt as supervision to train the model.
\subsection{Main Results}
\begin{table*}[h!]
	\centering
	\begin{tabular}{cccccc}
		\toprule
		\multirow{2}{*}{\textbf{Dataset}} & \multirow{2}{*}{\textbf{Labels}} & \multicolumn{3}{c}{\textbf{Relation}} & {\textbf{Overall}}\\ 
		& & $Pre.$ & $Rec.$ & $F_1$ & $Acc.$\\
		\midrule
		\multirow{3}{*}{Wiki-ZSL(single)} & m=5 & 81.54$\to$ 20.18 & 45.51$\to$50.35  & 58.41$\to$28.81 & 18.40$\to$ 16.71\\
        & m=10 & 55.08$\to$9.89 & 57.75$\to$49.56 & 56.38$\to$16.49 & 22.30$\to$15.52 \\
        & m=15 & 49.64$\to$6.69 & 56.35$\to$50.15 & 52.78$\to$11.80 & 21.64$\to$15.92 \\
        \midrule
        \multirow{3}{*}{FewRel(single)} & m=5 & 89.40$\to$20.19 & 54.80$\to$50.51 & 67.95$\to$28.85 & 22.67$\to$19.67 \\
        & m=10 & 69.65$\to$10.02 & 59.64$\to$50.18 & 64.26$\to$16.70 & 24.91$\to$19.25 \\
        & m=15 & 55.66$\to$6.67 & 64.61$\to$49.97 & 59.81$\to$11.77 & 25.14$\to$17.51 \\
        \midrule
        & & $Pre.$ & $Rec.$ & $F_1$ & $F_1$\\
        \midrule
        \multirow{3}{*}{Wiki-ZSL(multi)} & m=5 & 93.39$\to$19.67 & 69.55$\to$47.93 & 79.72$\to$ 27.89 & 37.48$\to$16.60\\
        & m=10 & 76.15$\to$12.33 & 68.23$\to$50.47 & 71.97$\to$19.82 & 32.00$\to$10.72 \\
        & m=15 & 61.68$\to$7.83 & 61.74$\to$49.87 & 61.71$\to$13.53 & 28.98$\to$7.63 \\
        \midrule
        \multirow{3}{*}{FewRel(multi)} & m=5 & 94.65$\to$20.00 & 66.67$\to$51.80 & 78.23$\to$28.86 & 38.93$\to$16.87 \\
        & m=10 & 78.46$\to$11.29 & 64.65$\to$50.92 & 70.89$\to$18.84 & 30.39$\to$10.42 \\
        & m=15 & 65.33$\to$7.62 & 53.38$\to$49.00 & 58.76$\to$13.19 & 25.16$\to$7.47 \\
		\bottomrule
	\end{tabular}
	\caption{The model performance after replacing potential candidate relation selection with random guess}
	\label{tab:further}
\end{table*}
The model performance on two ZeroRTE datasets is reported in Table~\ref{tab:result}.
The results show that PCRED consistently outperforms previous methods.
For single triplet extraction, PCRED outperforms RelationPrompt on both Wiki-ZSL and FewRel, with performance gains in $Acc.$ ranging from 0.40\% to 6.17\%.
For multiple triplet extraction, PCRED also outperforms RelationPrompt on Wiki-ZSL and FewRel, with performance gains in the $F_1$ score ranging from 0.13\% to 16.59\%.
Although PCRED does not consistently achieve the highest precision or recall scores, the precision and recall scores are balanced, and hence it reaches the highest $F_1$ score.
Other baseline methods like TableSequence or RelationPrompt (NoGen) have either imbalanced precision or recall scores, leading to lower $F_1$ scores.

Moreover, PCRED does not rely on any additional training samples, while RelationPrompt is severely  dependent on the synthetic data.
Compared with RelationPrompt (NoGen) which does not use synthetic data, performance gains of RCRED are much more significant.
For single triplet extraction, RCRED outperforms RelationPrompt (NoGen) from 9.35\% to 15.20\%;
for multiple triplet extraction, PCRED achieves performance gains in the $F_1$ score ranging from 15.22\% to 24.36\%.
It indicates that RelationPrompt does not fully exploit the potential zero-shot ability of PLMs, while PCRED achieves better performance by leveraging the semantics of relations to recognize unseen relations.
It is worth noting that PCRED has only 56.8\% of the model parameters against to RelationPrompt (150M against 264M).
These results further demonstrate the simplicity and effectiveness of our proposed method.

\subsection{Further Analysis}
The core module of PCRED is the potential candidate relation selection module that can recognize unseen relation labels based on the semantics of relations.
To further demonstrate the effectiveness of this module, we conduct further analysis.
We replace the potential candidate relation selection module with a random predictor, which predicts potential relations by random guess. 
The output probability of each candidate relation conforms to a uniform distribution between 0 and 1.
The result is reported in Table~\ref{tab:further}.

As the result shows, the precision score of relation selection is remarkably higher than that of random guess, which indicates that PCRED can recognize unseen relations based on the semantics instead of random guess.
Besides, most of the relation recall is above 50\%
, leading to a more balanced relation $F_1$ score.
After we replace the potential candidate relation selection module with a random predictor, the overall performance shows a drastic decline, especially for multiple triplet extraction, with a performance drop for at most 22.06\%.

\subsection{Visualization}
\begin{figure}
	\centering
	\includegraphics[width=\linewidth]{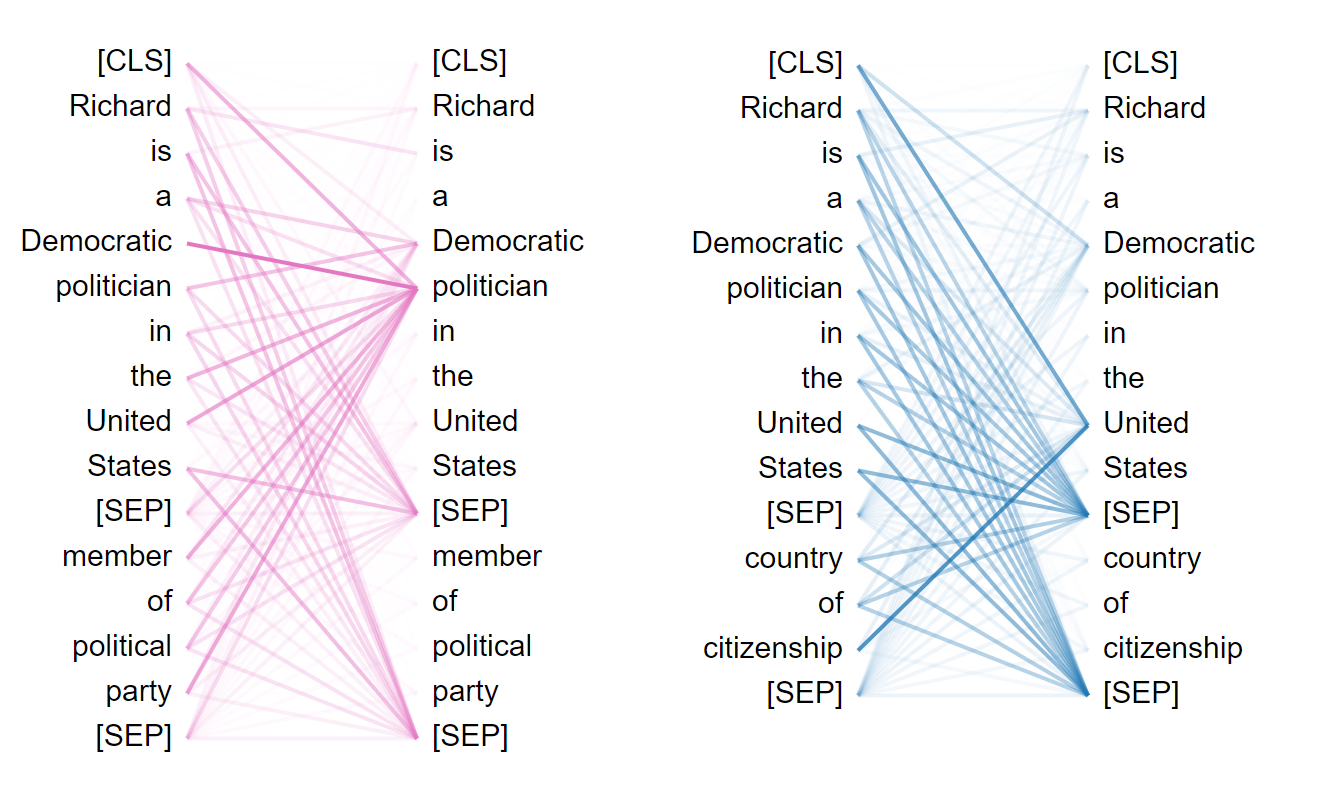}
	\caption{The visualization of the self-attention matrix}
	\label{fig:attention}
\end{figure}

We use Bertviz~\cite{vig-2019-multiscale} to visualize the self-attention matrix from the last layer of our model in Figure~\ref{fig:attention} to show that concatenating relations as additional context does infuse the semantics of relations.
As the figure demonstrates, when we concatenate the relation \textit{member of political party} to the original sentence, the [CLS] token pays more attention to tokens \textit{Democratic} and \textit{politician}; when we concatenate the relation \textit{country of citizenship} to the sentence, the [CLS] token concentrates on the token \textit{United}. 
It indicates that different relations indeed prompt the [CLS] token to focus on different context information, which further demonstrates the effectiveness of our proposed relation selection method in recognizing unseen relations.
\section{Related Work}
\subsection{Relation-first relation triplet extraction}
Even though traditional RET methods cannot solve the ZeroRTE task, some of them can still provoke new thinking for ZeroRTE.
PCRED adopts a relation-first paradigm, which eliminates the need for additional training data. 
The relation-first paradigm can be traced back to~\citeposs{takanobu2019hierarchical} work, which applies a hierarchical reinforcement framework, and detects relations first with a high-level reinforcement learning process.
\citet{ma2021effective} provide a theoretical analysis to prove that the relation-first paradigm is mathematically better than the entity-first paradigm.
Previous state-of-the-art traditional RTE method PRGC~\cite{zheng-etal-2021-prgc} decomposes the task into three sub-tasks, and the first sub-task is relation judgement.
However, these methods only treat relations as discrete class IDs or learnable embeddings, while the semantics of relations are ignored.

To better leverage the semantics of relations, \citet{li-etal-2022-rfbfn} manually define a question template for each relation. 
Their model RFBFN first detects potential relations.
Each relation has a corresponding manually-designed relation template with blanks.
The entity extraction is performed by filling the blanks.
However, RFBFN still cannot be applied to ZeroRTE, since human efforts are required to design the question template. 
Moreover, it is unrealistic to design templates for unseen relations. 
Unlike RFBFN, we directly concatenate the text of relations to the original sentence, which can not only infuse the semantics of relations in the context but save the burden of designing templates as well. 
\subsection{Zero-shot relation extraction}
Although there are existing studies on zero-shot relation extraction,
they do not require the extraction of whole triplets.
\citet{levy-etal-2017-zero} perform zero-shot relation extraction via reading comprehension.
They reduce the task into a slot-filling problem by manually-designed question templates.
However, the model only predicts the tail entity based on the provided head entity and relations.
\citet{chen-li-2021-zs} propose a model named ZS-BERT for zero-shot relation classification with attribute representation learning. ZS-BERT encodes sentences and the description of relations into a shared space. The prediction is obtained by nearest neighbor search.
However, ZS-BERT can only infer relations and assume that the ground-truth entity pairs are readily available, which is unrealistic in real scenarios.

RelationPrompt is the first approach that can extract the whole triplet under the zero-shot setting.
As mentioned before, its performance is severally constrained by synthetic data. 
Besides, it is hard to control the quality of the synthetic data generated by PLMs. It remains to explore how to better leverage PLMs to tackle the challenging ZeroRTE task.
\section{Conclusion}
We propose a novel method PCRED with potential candidate relation selection and entity boundary detection.
Instead of leveraging PLMs to generate training samples of unseen relations, our model directly utilizes the semantics of unseen relations by concatenating them to the sentence as additional context.
The experiment results on two ZeroRTE datasets show that our method consistently outperforms previous works.
Further analysis and the visualization of the attention matrix prove the effectiveness of the potential candidate relation selection.
\section*{Limitations}
We only conduct experiments on two sentence-level ZeroRTE datasets, and it remains to explore the model performance on document-level ZeroRTE datasets. 
It is much more challenging to extend the method to document-level ZeroRTE, since one entity may contain multiple mentions, which cannot be merely determined by a single boundary. Nevertheless, the idea of potential candidate relation selection still works.

\bibliography{anthology,custom}
\bibliographystyle{acl_natbib}
\appendix
\section{Appendix}
\subsection{Implementation Details}
\label{sec:appendix}
We implement our works under the Pytorch~\cite{paszke2019pytorch} and Transformers~\cite{wolf2019huggingface} frameworks. 
We present all the hyper-parameters in Table~\ref{tab:param} for reproduction.
All the experiments are conducted on a Nvidia-GeForce RTX-3090 GPU.
We employ the base version of BERT, and freeze its pretrained word embeddings during training.
The batch size is set to 16 and the maximum training epoch is 10.
The learning rate is set to 0.00005 with a linear warm-up strategy for the 20\% of the total training steps.
AdamW~\cite{loshchilov2018decoupled} is used as the optimizer. 
We apply an early stopping strategy based on the validation score.
If the score does not increase for up to 4 epochs, the model will stop training.
We save the best checkpoint on the validation set and load it for testing.
We fix the random seed so that the result is deterministic.
\begin{table}[!h]
	\centering
	\begin{tabular}{cc}
        \hline
		\textbf{Hyper-parameter} & \textbf{Value} \\
		\hline 
		batch-size & 16 \\
		max-epoch & 10 \\
		learning-rate & 0.00005 \\
		early-stopping-patience & 4 \\
		max-span-length & 15 \\
		max-sequence-length & 100 \\
		relation-threshold ($\delta$) & 0.5 \\
		boundary-threshold ($\beta$) & 0.4 \\
		\multirow{2}{*}{group-size ($\mathcal{G}$)}  & 5 (FewRel) \\
		& 6 (Wiki-ZSL) \\
		\multirow{2}{*}{max-triplets ($\mathcal{N}$)} & 4 (FewRel) \\
		& 6 (Wiki-ZSL) \\
		warm-up-ratio & 0.2 \\
		loss-weight ($\alpha$) & 1 \\
		\hline
	\end{tabular}
	\caption{All the hyper-parameters of the experiment}
	\label{tab:param}
\end{table}
\end{document}